\documentclass[conference]{IEEEtran}
\IEEEoverridecommandlockouts
\usepackage{cite}
\usepackage{amsmath,amssymb,amsfonts}
\usepackage{algorithmic}
\usepackage{graphicx}
\usepackage{textcomp}
\usepackage{xcolor}
\def\BibTeX{{\rm B\kern-.05em{\sc i\kern-.025em b}\kern-.08em
    T\kern-.1667em\lower.7ex\hbox{E}\kern-.125emX}}

\usepackage{xspace}
\usepackage{mathtools}
\usepackage{booktabs}
\usepackage{multirow}
\usepackage[ruled, linesnumbered]{algorithm2e}
\usepackage{stfloats}
\usepackage{subcaption}
\usepackage{bm}
\usepackage{hyperref}
\usepackage{makecell}

\newcommand{\norm}[1]{\left\lVert#1\right\rVert}
\DeclareMathOperator*{\argmin}{arg\,min}
\DeclareMathOperator*{\argmax}{arg\,max}

\makeatletter
\newcommand{\removelatexerror}{\let\@latex@error\@gobble}
\makeatother

\newcommand{\knn}[0]{kNN\xspace}

\newcommand{\papernot}[0]{Papernot \& McDaniel\xspace}
\newcommand{\swa}[0]{Sitawarin \& Wagner\xspace}

\begin{document}

\title{Minimum-Norm Adversarial Examples on \\ KNN and KNN-Based Models}

\author{
\IEEEauthorblockN{Chawin Sitawarin}
\IEEEauthorblockA{EECS Department, UC Berkeley\\chawins@eecs.berkeley.edu}
\and
\IEEEauthorblockN{David Wagner}
\IEEEauthorblockA{EECS Department, UC Berkeley\\daw@cs.berkeley.edu}
}

\maketitle

\begin{abstract}
We study the robustness against adversarial examples of \knn classifiers and classifiers that combine \knn with neural networks. The main difficulty lies in the fact that finding an optimal attack on \knn is intractable for typical datasets. In this work, we propose a gradient-based attack on \knn and \knn-based defenses, inspired by the previous work by \swa \cite{sitawarin19dknn}. We demonstrate that our attack outperforms their method on all of the models we tested with only a minimal increase in the computation time. The attack also beats the state-of-the-art attack \cite{yang19nonpara} on \knn when $\bm{k > 1}$ using less than 1\% of its running time. We hope that this attack can be used as a new baseline for evaluating the robustness of \knn and its variants.
\end{abstract}

\section{Introduction}

Adversarial examples, ordinary samples that are slightly perturbed to fool machine learning models, highlight the fragility of neural networks \cite{biggio13,szegedy13,goodfellow14explaining,moosavi15deepfool,nguyen15}. Recently, this has motivated research on classical \knn classifiers to study their robustness properties and to leverage \knn for providing interpretability and robustness to neural networks \cite{papernot18dknn,wang18knn,sitawarin19def,dubey19webnn}.

Accurately evaluating the robustness of large and complex models is a challenging problem \cite{carlini16distill,carlini17bypass,athalye18}. Evaluating \knn-based models is particularly difficult since they are not differentiable: most existing attacks rely on gradient descent and thus can't be applied to \knn-based models. Yang et al. propose a dedicated attack on \knn  \cite{yang19nonpara}, but it scales poorly with $k$, the number of data points, and/or the dimension. \swa propose a heuristic for solving this problem by coming up with a differentiable loss function for the adversary and then relying on gradient descent to adversarial examples~\cite{sitawarin19dknn}.

In this work, we propose a gradient-based method of finding adversarial examples that improves on \swa. We find that their attack was not optimal, and in all of experiments, our method outperforms it by a large margin, finding adversarial examples with smaller perturbation. Our method also improves on Yang et al.: when $k>1$, we find adversarial examples with smaller perturbation, using less than 1\% of the running time.  Thus, our attack represents the new state of the art for attacking \knn classifiers.

We are also interested in evaluating the robustness of several defenses that apply \knn to the intermediate outputs from some layers of a neural network. Since our attack relies on gradient descent, it can be readily applied to these defenses, whereas Yang et al. cannot. We apply our attack to Deep \knn \cite{papernot18dknn}, single-layered Deep \knn \cite{sitawarin19def}, and a defense by Dubey et al. \cite{dubey19webnn}, and use it to evaluate the security of those defenses. We find that our method outperforms the original algorithm by \swa in both the success rate and the size of the perturbation. We find that all of the mentioned approaches can appear falsely robust when a weaker attack is used during their evaluation. After applying our improved attack, none of them appear to be more robust than an adversarially trained network \cite{madry17}.

Similar to all of the gradient-based attacks, our approach is fast, but it does not provide a certificate of robustness or a guarantee that no smaller adversarial perturbation does not exist. Nonetheless, it serves as an efficient and reliable baseline for evaluating \knn-based models\footnote{Please find our code at \url{https://github.com/chawins/knn-defense}}.

\section{Background and Related Work}

\subsection{Adversarial Examples}

Adversarial examples are inputs specifically crafted to induce an erroneous behavior from machine learning models, usually by adding a small imperceptible perturbation. The robustness property of machine learning classifiers has been studied for a long time both in the random noise and in the adversarial settings \cite{barreno06can,huang11advml}. The field gained attention after adversarial perturbations were shown on deep neural networks \cite{szegedy13,goodfellow14explaining}. In general, finding such a perturbation can be formulated as a constrained optimization problem:
\begin{align*}
    x_{adv} = x + \delta^* \quad \text{where} ~ \delta^* = ~ &\argmax_{\delta} ~~ L(x + \delta, y) \\
    &\text{such that} ~ \norm{\delta}_p \leq d \nonumber
\end{align*}
where $L(\cdot)$ is some loss function, $x$ a clean sample, and $y$ its associated true label. The constraint is used to keep the perturbation small or \textit{imperceptible} to humans. The difficulty of solving this problem depends on the loss function and on the machine learning model. For neural networks, this problem is non-convex.  In our case where the model involves some form of \knn, the loss function is not smooth, and naive methods to solve the optimization problem need to iterate through every possible set of $k$ neighbors, which is intractable for large $k$.

\subsection{KNN-Based Defenses}

The \knn classifier is a popular non-parametric classifier that predicts the label of an input by finding its $k$ nearest neighbors in some distance metric (such as Euclidean or cosine distance) and taking a majority vote from the labels of those neighbors. 

The robustness of \knn is well-studied in the context of label noise \cite{gao16,reeve19}. Recently, multiple papers have presented evidence of its robustness in adversarial settings from both theoretical perspectives \cite{wang18knn,khoury19voronoi} and empirical analyses \cite{papernot16transfer,papernot18dknn,schott18towards,dubey19webnn}. Despite its potential, \knn is known to struggle on high-dimensional data like most of real-world datasets. Hence, a few works attempt to combine \knn with more complex feature extraction in order to obtain the robustness benefits of \knn while maintaining good accuracy. In this work, we re-evaluate the robustness of \knn and other \knn-based models that use neural networks as a feature extractor.

We examine three schemes. Deep \knn \cite{papernot18dknn} uses a nearest neighbor search on each of the deep representation layers and makes a final prediction based on the sum of the votes across all layers. Dubey et al. concatenates the representations from all layers, applies average pooling and dimensionality reduction using PCA, and then applies \knn \cite{dubey19webnn}; the final prediction is a weighted average of the logits of the $k$ nearest neighbors in the training set. Here, we only consider their uniform weighting scheme. Single-layered Deep \knn \cite{sitawarin19def} is similar to Deep \knn, but only uses the penultimate layer for the \knn search.  We also consider a variant where the network is adversarially trained to increase the robustness. We refer readers to the original works for more detailed descriptions.

\subsection{Existing Attacks on kNN}

Yang et al. \cite{yang19nonpara} propose a method for finding the smallest perturbation that changes the classification of a given \knn by solving a quadratic optimization problem. In short, they minimize the $\ell_2$-norm of the perturbation subject to a set of linear constraints that keep the adversarial example in a certain Voronoi cell. This formulation can find the smallest perturbation exactly and fairly efficiently for $k=1$, but it is intractable for $k>1$ since the optimization problem must be repeatedly solved over all Voronoi cells that are classified as a different label from the original, and the number of such Voronoi cells is exponential in $k$. The authors mitigate this problem with a heuristic that only considers a small set of $s'$ ``nearby'' Voronoi cells. We refer readers to the original paper for more details. 

Even then Yang et al.'s approach still suffers from two problems. First, solving many quadratic problems with a large number of constraints in high dimension is still too expensive for real-world datasets. Second, this method cannot be applied to defenses that combine a neural network and \knn, because the linear constraints in the representation space cannot be enforced by linear constraints on the input. 

\swa propose a heuristic gradient-based attack on \knn and Deep \knn that circumvents the two previous issues. However, when compared to the exact approach of Yang et al. on \knn, the adversarial examples found by \swa require a larger perturbation which can lead to an inaccurate robustness evaluation. In this work, we improve this attack significantly and show that it can be easily extended to attack the other \knn-based defenses with much smaller perturbations than previously reported.

\section{Threat Model}

We consider a complete white-box threat model for this work, meaning that the adversary has access to the training set, the value of $k$ and the distance metric used in \knn, and all parameters of the neural networks. We also only consider $\ell_2$ attacks as gradient-based approaches generally do not work well for finding minimum-$\ell_\infty$-norm adversarial examples. We only discuss robustness against an untargeted attack, but our method easily generalizes to targeted attacks.

\section{Attack on K-Nearest Neighbors} \label{sec: attack_knn}

\subsection{Notation}

Let $x \in \mathbb{R}^d$ denote a target sample or a clean sample that the adversary uses as a starting point to generate an adversarial example, and $y \in \{1,2,...,c\}$ its ground-truth label. We denote the perturbed version of $x$ as $\hat{x} = x + \delta$. The training set used by \knn and to train the neural network is $(X, Y)$ where $X = \{x_1,x_2,...,x_n\}$ and $Y = \{y_1,y_2,...,y_n\}$. 

Here, we only consider \knn that uses Euclidean distance as the metric. \knn returns an ordering of the indices of the neighbors of an input $x$ from the nearest to the $k$-th nearest one (i.e., $\pi_1(x),...,\pi_k(x)$). The $k$ nearest neighbors of $x$ are then given by $x_{\pi_1(x)},...,x_{\pi_k(x)}$, and it follows that
\begin{align*}
    \norm{x - x_{\pi_i(x)}}_2 < \norm{x - x_{\pi_j(x)}}_2 \iff i < j
\end{align*}
The final prediction of a \knn can be written with a majority function $\text{Maj}(\cdot)$ as follows:
\begin{align*}
    \mathrm{knn}(x) = \text{Maj}(y_{\pi_1(x)},...,y_{\pi_k(x)})
\end{align*}

\subsection{\swa Attack}

First, we summarize the attack proposed by \swa, and in the next section, we will describe our improvements. \swa solve the following optimization problem:
\begin{align*}
    \hat{\delta} ~ = &\argmin_\delta ~ \sum_{i=1}^{m} ~w_i \cdot \sigma \left( \norm{\tilde{x}_i - (x + \delta)}_2^2 - \eta^2 \right) + c \norm{\delta}_2^2\\
    &\text{such that} ~ x + \delta \in [0, 1]^d \nonumber
\end{align*}
where $\sigma(\cdot)$ is a sigmoid function and $\eta$ is a threshold initially set to the $\ell_2$-distance between $x$ and $x_{\pi_k(x)}$ (i.e. $\eta = \norm{x - x_{\pi_k(x)}}_2$). The second term in the objective penalizes the norm of the perturbation, and the balancing constant $c$ is obtained through a binary search \cite{carlini17cw}. 

The constraint ensures that the adversarial example lies in a valid input range, which here we assume to be $[0, 1]$ for MNIST pixel values. Similarly to Carlini \& Wagner attack, this constraint is enforced by using a change of variable (e.g. $\tanh$ function). This constraint is implicitly assumed and will be omitted for the rest of the paper. The optimization problem can then be solved with any choice of gradient descent algorithm. The original paper chooses the popular Adam optimizer.

The attack uses a subset of the training samples called \textit{guide samples}, denoted $\{(\tilde{x}_1, \tilde{y}_1),...,(\tilde{x}_m, \tilde{y}_m)\}$. These can be chosen by different heuristics. The heuristic empirically found by \swa to work best is to choose a set of $m$ training samples all from the same class not equal to $y$ with the smallest mean distance to $x$, i.e., the guide samples are given by
\begin{align*}
    \argmin_{(\tilde{X}, \tilde{Y}) \subset (X, Y), ~ |(\tilde{X}, \tilde{Y})|=m, ~ y_{adv} \ne y} ~ &\sum_{\tilde{x} \in \tilde{X}} \norm{\tilde{x} - x}_2 \\
    \text{such that} \qquad &\forall \tilde{y} \in \tilde{Y}, ~ \tilde{y} = y_{adv}
\end{align*}
Guide samples can be found by minimizing this objective for each label $y_{adv} \ne y$ and then picking the smallest one. We fine-tune the hyperparameter $m$. The coefficients $w_i$ are set to 1 if $\tilde{y}_i \ne y$, otherwise $w_i = -1$. For this heuristic, $w_i = 1$ for all $i$'s. Informally, the \swa attack finds a perturbation so that the distance from $\hat{x}$ to each guide sample is less than some fixed threshold $\eta$.

\subsection{Our Attack Description}

\begin{figure}[!t]
\removelatexerror
\begin{algorithm}[H]
 \SetKwInOut{Input}{Input}
 \SetKwInOut{Output}{Output}
 \SetKwInOut{Parameters}{Parameters}
 \Input{Target sample and label $x,y$ \\
        Training samples $X,Y$ \\
        Number of neighbors in \knn $k$}
 \Parameters{$m, p, q$}
 \Output{Adversarial examples $x_{adv}$}
 Initialize $\hat{\delta}$ as some large vector \\
 \For{i = 1,\dots,q + 1}{
    \eIf{i = 1}{
        Initialize $\delta = 0$
    }{
        Initialize $x + \delta$ to $(i-1)$-th nearest sample of a class other than $y$ (see \textit{Attack Initialization})
    }
    Add random noise: $\delta \leftarrow \delta + \alpha, \alpha \sim \mathcal{N}(0, 0.01)$ \\
    \For{j = 1,\dots,MAX\_STEPS}{
        \If{j $\bmod$ p = 0}{
            Update $m$ guide samples $(\tilde{X}, \tilde{Y})$ and threshold $\eta$ (see \textit{Dynamic Threshold and Guide Samples} \& \textit{Heuristic for Picking Guide Samples})
        }
        Take a gradient step on Eq. \ref{eq:knn_attack} to update $\delta$
    }
    Repeat lines 9--14 with a binary search on the constant $c$ \\
    \If{$\mathrm{knn}(x + \delta) \ne y$ and $||\delta||_2 < ||\hat{\delta}||_2$}{$\hat{\delta} \leftarrow \delta$}
 }
 \Return{$x_{adv} = x + \hat{\delta}$}
 \caption{Our Attack on \knn}
 \label{alg}
\end{algorithm}
\end{figure}

Our attack is inspired by the idea and the formulation of \swa. Our algorithm solves the following optimization problem:
\begin{align} \label{eq:knn_attack}
    \hat{\delta} = \argmin_\delta ~ &\sum_{i=1}^{m} ~ \max \left\{w_i\left(\norm{\tilde{x}_i - (x + \delta)}_2^2 - \eta^2 \right) + \Delta, 0 \right\} \nonumber \\ 
    &+ c\norm{\delta}_2^2
\end{align}
The changes compared to the original version are to use ReLU instead of sigmoid and the introduction of $\Delta$. 

\textbf{Threshold function:} The original idea of using the sigmoid is to simulate \knn's hard threshold with a soft differentiable one. The goal is to put the perturbed sample ``just a bit further'' from the training samples of the correct class than the threshold and ``just a bit closer'' to ones of a different class than the threshold, which is the distance between the input and its $k$-th nearest neighbor. 
We propose that ReLU can achieve a similar effect to sigmoid, and it avoids the need to deal with overflow and underflow problems caused by the exponential in the sigmoid function. \swa uses a sigmoid because it targets \knn that uses cosine distance, and hence, the distance always lies between $-1$ and $1$, but handling the Euclidean distance is challenging with this approach.  ReLU avoids these problems.

\textbf{Creating a small gap:} $\Delta$ is added for numerical stability. We do not want $\hat{x}$ to be at exactly the same distance from all guide samples (both from the correct and the incorrect classes) as we have to deal with a tie-breaking. Choosing $\Delta$ as any small positive constant such as $1 \times 10^{-5}$ ensures that $\hat{x}$ is a bit closer to the guide samples from the incorrect class than the ones from the correct class. 

In addition to the change to the objective function, we also introduce a number of heuristic improvements (summarized in Algorithm \ref{alg} and explained below):

\subsubsection{Dynamic Threshold and Guide Samples}
\swa fixed the threshold $\eta$ and the set of guide samples at initialization and never changed them afterwards. Unfortunately, as $\hat{x}$ moves further and further from $x$, they become less suitable. For instance, the distance from $\hat{x}$ to its $k$-th nearest neighbor changes with $\delta$, i.e. the original $\eta(x) = \norm{x - x_{\pi_k(x)}}_2$ is no longer a good approximation of $\eta(x + \delta) = \norm{x + \delta - x_{\pi_k(x + \delta)}}_2$. Also, $\hat{x}$ may get closer to other training samples of the correct class that were not initially one of the guide samples. Each of these can cause the objective function to be a poor approximation to \knn's decision rule. We address this problem by dynamically recomputing $\eta(\hat{x})$ and the guide samples every $p$ steps of gradient descent. Ideally, they could be recomputed at every step (i.e. $p = 1$), but doing so is very computationally expensive. As a result, $p$ should be chosen as small as possible ($\sim10-100$) until the benefit of lowering $p$ is negligible or until the running time becomes too large.

\subsubsection{Heuristic for Picking Guide Samples}
We experiment with different heuristics for choosing the guide samples and find it is better to pick half of them ($m/2$ instead of $m$) in the same manner as the original and the other half as the $m/2$ nearest training samples of the correct label:\\
$\{(\tilde{x}_1, \tilde{y}_1),...,(\tilde{x}_{m/2}, \tilde{y}_{m/2})\} = $
\begin{align*}
    \argmin_{(\tilde{X}, \tilde{Y}) \subset (X, Y), ~ |(\tilde{X}, \tilde{Y})|=m/2, ~ y_{adv} \ne y} ~ &\sum_{\tilde{x} \in \tilde{X}} \norm{\tilde{x} - x}_2 \\
    \text{such that} \qquad &\forall \tilde{y} \in \tilde{Y}, ~ \tilde{y} = y_{adv}
\end{align*}
$\{(\tilde{x}_{m/2 + 1}, \tilde{y}_{m/2 + 1}),...,(\tilde{x}_{m}, \tilde{y}_m)\} = $
\begin{align*}
    \argmin_{(\tilde{X}, \tilde{Y}) \subset (X, Y), ~ |(\tilde{X}, \tilde{Y})|=m/2} ~ &\sum_{\tilde{x} \in \tilde{X}} \norm{\tilde{x} - x}_2 \\
    \text{such that} \qquad &\forall \tilde{y} \in \tilde{Y}, ~ \tilde{y} = y
\end{align*}
The coefficients $w_i$ are picked in the same manner as the original. We suggest that $m$ should be chosen as small as possible, starting from $k$ if $k$ is even or $k+1$, otherwise. A smaller $m$ means the optimization takes into account fewer guide samples, which results in a shorter running time and generally smaller perturbation. Hence, $m$ should be increased only when the current attack is unsuccessful.

In a targeted attack, one can simply choose $y_{adv}$ as the target label, and everything remains the same. A strictly stronger version of our untargeted attack is to perform a targeted attack $c-1$ times, once for each possible class.  In other words, (1) pick guide samples with a fixed label $y_{adv} \ne y$, (2) solve the optimization problem, then (3) repeat the process again for every possible $y_{adv}$, and keep the adversarial examples with the smallest perturbation. However, doing so will increase the running time by a factor of $c - 1$. 

\subsubsection{Attack Initialization}
Attacks on \knn-based defenses can get stuck in a local minimum and fail to find an adversarial example. Adding random noise at initialization does not completely solve the problem. Since the neural network part makes the optimization non-convex, good initialization is important. This problem can be regarded as a mild instance of the gradient obfuscation problem \cite{athalye18}. To guarantee the success of the attack and circumvent this problem, we initialize $\hat{x}$ as a nearby sample from one of the incorrect classes. More specifically, we find the $q$ nearest training samples to $x$ that are classified as any class other than $y$, and then run the attack $q$ times, each time starting from one of those samples. The larger $q$ is, the stronger our attack, but it also increases the running time of the attack.

\subsubsection{Other Minor Changes}
On top of the first three major changes, we also add a few techniques to further enhance the attack. First, we slowly decrease $m$ if the attack succeeds; otherwise, we keep it the same. Assuming that the first term in Eq. \ref{eq:knn_attack} can be reduced zero, choosing $m = k$ (or $k+1$) guarantees that the attack succeeds. Nonetheless, the optimization sometimes can get stuck in local minima when combined with a neural network and fails to find an adversarial example. This can be mitigated by using $m > k$ (or $k + 1$), but doing so could lead to a larger-than-optimal perturbation. Hence, we implement an automatic way of reducing $m$ if possible until it reaches $k$ (or $k + 1$).

Second, we periodically classify the perturbed sample every certain number of optimization steps. If the perturbed sample is misclassified and has the smallest perturbation encountered so far, we save it. This helps preventing the optimization from overshooting and finding a larger perturbation that it needs to be. Finally, we found that RMSprop worked slightly better than the Adam optimizer.

\section{Attacks on KNN-Based Defenses}

Our attack can be applied to models based on \knn with very little modification. In this section, we briefly describe how to adapt our attack for \knn to the three models that combine \knn with neural networks.

\subsection{Deep kNN}
We attack the kNNs on all layers simultaneously, and the loss is summed across the layers before backpropagation. In other words, the optimization objective in Eq. \ref{eq:knn_attack} can be rewritten as follows:
\begin{align*}
    \sum_{l \in L} \sum_{i=1}^{m} ~ \max \left\{w_i\left(d_l(\tilde{x}_i, x+ \delta)^2 - \eta_l^2\right) + \Delta, 0 \right\} + c\norm{\delta}_2^2 \\
    \text{where } ~ d_l(x_1, x_2) = \norm{f_l(x_1) - f_l(x_2)}_2
\end{align*}
where $d_l(\cdot, \cdot)$ is Euclidean distance in the representation space output by a neural network at layer $l$, and $L$ is a set of such layers used by the \knn. Solving this objective can be computationally intensive if the features are high-dimensional. The guide samples are computed based on only the first layer and used for all subsequent layers, but the threshold still must be calculated separately for each layer.

\subsection{Single-Layered Deep \knn}
The attack on deep \knn can be directly applied to single-layered deep \knn: here $L$ only contains the penultimate layer of the neural network.

\subsection{Dubey et al.}
Dubey et al. do not use \knn directly on intermediate representations of the network. Instead, \knn is applied on a linearly transformed space. The distance metric must then be computed in this space:
\begin{align*}
    d(x_1, x_2) = \norm{A(\phi(x_1) - \mu) - A(\phi(x_2) - \mu)}_2 
\end{align*}
where $\phi(x)$ is a concatenation of an average pooling of $f_l(x)$ for every layer $l$ used by the model, $A$ is a PCA transformation matrix that reduces dimension of the features, and $\mu$ is the mean of $\phi(x)$ across the training set: $\mu = \frac{1}{n} \sum_{i=1}^{n} \phi(x_i)$.

\section{Experimental Setup}

We evaluate our new attack by comparing it to \swa and Yang et al (Section \ref{ss:knn}). Duplicating the same setting in Yang et al., we use two-class MNIST (3 vs 5) with 1000 training samples from each class. To highlight the scalability of our approach, we use normal-sized MNIST ($28 \times 28$ pixels) instead of reducing its dimension to 25 by PCA as in Yang et al. We use Yang et al.'s implementation of their attack\footnote{\url{https://github.com/yangarbiter/adversarial-nonparametrics}}. All  experiments are run on an Intel(R) Core(TM) i7-6850K CPU (3.60GHz) and one Nvidia 1080 Ti GPU.

Then, we evaluate several \knn-based defenses using our attack (\S~\ref{ss:dknn}).  Each defense used the same neural network architecture as in \papernot and \swa (three convolution layers followed by a final fully-connected layer). We evaluate deep \knn, single-layered deep \knn, and Dubey et al. Additionally, we also evaluate all of the above models with an adversarially trained version of the same network. All networks are trained on the full MNIST training set with a batch size of 128, learning rate of $1\times10^{-3}$, and an Adam optimizer. The adversarial training \cite{madry17} uses 40-step PGD with a step size of 0.2 and $\epsilon$ of 3.

Our deep \knn model uses cosine distance and $k$ is set to 75 for all of the four layers. We re-implemented Dubey et al.'s scheme and trained it on MNIST, passing features from the three convolutional layers (before ReLU) through an average pooling followed by PCA to reduce the dimension to 64. The \knn is then used on this 64-dimension representation, with $k=50$. Lastly, the single-layered deep \knn uses $k=5$ with \knn applied to the representation output by the last convolutional layer.

We measure the performance of our attack and the robustness of these defenses by the mean $\ell_2$-norm of the adversarial perturbation. An attack is considered better if it finds a smaller perturbation that fools the target model. For \knn, we evaluate all of the attacks on 200 test samples (100 samples from each class) regardless of their original classification results, similarly to Yang et al. For the \knn-based defenses, adversarial examples are generated for the first 100 samples in the test set that are correctly classified originally by each respective model.

\section{Results} \label{sec: results}

\subsection{Comparing Attacks on kNN} \label{ss:knn}

\begin{table}[t]
\centering
\caption{Comparison of attacks on \knn.}
\label{tab:knn}
\begin{tabular}{@{}clrr@{}}
\toprule
$k$                & Attacks               & Mean $\ell_2$-Norm & Approx. Running Time \\ \midrule
\multirow{4}{*}{1} & Yang et al. (Exact)   & \textbf{2.4753} & 30 hrs \\
                   & Yang et al. (Approx.) & \textbf{2.4753} & 2 hrs \\
                   & \swa   & 3.4337 & 1 mins \\
                   & Ours   & 2.7475 & 5 mins \\ \midrule
\multirow{3}{*}{3} & Yang et al. (Approx.) & 2.9857 & 11 hrs \\
                   & \swa   & 3.9132 & 1 mins \\
                   & Ours   & \textbf{2.9671} & 5 mins \\ \midrule
\multirow{3}{*}{5} & Yang et al. (Approx.) & 3.2473 & 44 hrs \\
                   & \swa   & 3.9757 & 1 mins          \\
                   & Ours   & \textbf{3.0913} & 5 mins \\ \bottomrule
\end{tabular}
\end{table}

Table \ref{tab:knn} compares three \knn attack methods: Yang et al., \swa, and ours. For $k=1$, both the exact and the approximate versions ($s'=50$) of Yang et al. find adversarial examples with the same minimal perturbation. Our attack finds adversarial examples with about 10\% larger mean perturbation but using only about 5 minutes compared to 2 hours for Yang et al.'s approximate method.

For all values of $k$, our attack also find adversarial examples with smaller mean perturbation norm than \swa. The increase in our computation time compared to \swa mostly comes from recomputing the threshold and the guide samples as well as repeating the optimization with different initializations. 

Since the exact method from Yang et al. cannot scale to $k > 1$, we are left with the approximate version which still takes many hours to finish. Our attack outperforms their approximate attack for $k = 3, 5$, finding adversarial examples with smaller distortion and with significantly shorter computation time. Note that the running time reported above is a total over 100 samples. Since our attack mainly relies on gradient descent, it can be implemented with automatic differentiation software (e.g. PyTorch and Tensorflow) and run on GPUs. Hence, it fully benefits from parallelization, and the running time can be further improved by a factor of $q$ by also parallelizing different initializations. In this section, for all of our attacks, the parameters are set as follows: $m = k + 1, p = 20$ and $q = 3$.

\subsection{A Brief Runtime Complexity Analysis}

We argue that Yang et al.'s method scales poorly with $k$ in both the performance and the running time. First, the size of the perturbation found by their approximate method will be increasingly worse than the minimal perturbation as $k$ gets larger. Their heuristic for picking $s'$ Voronoi cells will miss exponentially more cells as $k$ increases since the total number of Voronoi cells is exponential in $k$ (with the worst case being $\binom{n}{k}$). To find the optimal perturbation, $s'$ would need to be exponential in $k$, but that would make the algorithm intractable.

Second, even when $s'$ is fixed as in our experiments, the quadratic program becomes slower as $k$ increases due to the increase in the number of Voronoi edges and thus the number of linear inequality constraints. A Voronoi cell has at most $k(n - k)$ or $\mathcal{O}(k)$ edges for $n \gg k$, meaning the number of linear constraints scales linearly with $k$. However, the running time of quadratic programs is polynomial in the number of constraints and hence, polynomial in $k$.

In contrast, a larger value of $k$ slightly increases the running time of our algorithm, but the scaling is far better than Yang et al. The increase in computation time mainly comes from a larger number of guide samples, which scales linearly with $k$. 

\subsection{Evaluating the Robustness of kNN-Based Defenses} \label{ss:dknn}

\begin{table}[t]
\centering
\caption{Evaluation of all the \knn-based models}
\label{tab:dknn}
\begin{tabular}{@{}llll@{}}
\toprule
Models & Clean Acc. & Attacks & Mean $\ell_2$-Norm \\ \midrule
Normal network   & 0.9878  & CW   & 1.3970             \\
Adversarial training \cite{madry17}  & 0.9647   & CW      & 2.8263             \\ \midrule[1pt]
\multirow{2}{*}{Ordinary \knn ($k=5$)}  & \multirow{2}{*}{0.9671} & SW   & 3.3185 (0.97) \\
                                        &                         & Ours & \textbf{2.1869} \\ \midrule
\multirow{2}{*}{Deep \knn (Normal) \cite{papernot18dknn}}        & \multirow{2}{*}{0.9877} & SW      & 1.9933             \\
                                                                          &                         & Ours    & \textbf{1.7539}             \\ \midrule
\multirow{2}{*}{Deep \knn (AT)}        & \multirow{2}{*}{0.9773} & SW      & 2.6792             \\
                                                                          &                         & Ours    & \textbf{2.4632}             \\ \midrule
\multirow{2}{*}{\makecell[cl]{Single-layered Deep \knn \\ (Normal) \cite{sitawarin19def}}} & \multirow{2}{*}{0.9910} & SW      & 2.0964 (0.94)      \\
                              &                         & Ours    & \textbf{1.5184}             \\ \midrule
\multirow{2}{*}{\makecell[cl]{Single-layered Deep \knn \\ (AT)}}             & \multirow{2}{*}{0.9871} & SW      & 2.9345 (0.97)      \\
                                                                          &                         & Ours    & \textbf{2.2992}             \\ \midrule
\multirow{2}{*}{Dubey et al. (Normal) \cite{dubey19webnn}}                          & \multirow{2}{*}{0.9838} & SW & 2.0057 \\                  
  & & Ours    & \textbf{1.2606}             \\ \midrule
\multirow{2}{*}{Dubey et al. (AT)}                         & \multirow{2}{*}{0.9680} & SW & 2.6985 \\                  
  & & Ours    & \textbf{2.3979}             \\
\bottomrule
\end{tabular}
\end{table}

We compare our attack with \swa (SW). As shown in Table \ref{tab:dknn}, our attack outperforms \swa on every model: it finds adversarial examples with smaller perturbation, and it successfully generate adversarial examples for all of the 100 test samples. In the table, ``normal'' means the model uses a normally trained network, and ``AT'' indicates that model uses an adversarially trained network. The numbers in the parentheses indicate the attack success rate if it is lower than 1, showing that \swa is not always successful at finding adversarial examples for the ordinary \knn and the two single-layered deep \knn models. 

As a reference, the first two rows above the thick black line report the robustness of the same neural network and its adversarially trained version used in the \knn-based defenses. We evaluate them with the popular CW attack \cite{carlini17cw}. This demonstrates that most of the defenses we evaluate are more robust than the normal neural network, but it also confirms that adversarial training is still a strong baseline that beats all of the defenses presented here. Without an appropriate attack or evaluation method, many models may falsely appear more robust than adversarial training when they are in fact not.

Using an adversarially trained network as a feature extractor for \knn is consistently more robust than using a normal network. However, this comes with a cost of clean accuracy. Thus, our experiments find a trade-off between robustness vs accuracy in this setting.

\section{Discussion}

\subsection{Deep \knn}
Deep \knn has the feature that it produces both a classification and a credibility score. The credibility score indicates its confidence in its prediction. We investigate whether our attack can generate adversarial examples with high credibility scores. Without any modification, our attack produces adversarial examples with very low credibility scores since it aims to find a minimum-norm perturbation. To increase the credibility score, we change the stopping condition of the optimization and save a perturbed sample only if it is misclassified with a large number of neighbors of the target class (hence, with high credibility). With all neighbors coming from the target class (credibility score of 1, the maximum), the mean perturbation norm is as large as 4.0323. With at least about 98\% of the neighbors coming from the target class (credibility score of 0.275 or larger), the mean perturbation norm is 3.3920. 

This size of perturbation is clearly visible to humans and contains semantic information of the target class. This result suggests that deep \knn's credibility score is potentially a good metric for filtering out $\ell_2$-adversarial examples on MNIST. Nonetheless, a more appropriate method of picking guide samples or a better attack may prove this statement wrong.

\subsection{Dubey et al.}
Our attack has a clear advantage over the ones used to evaluate Dubey et al.'s ImageNet model, because of our dynamic guide samples and the threshold function. Thus, it is hard to know whether Dubey et al.'s scheme truly improves robustness. However, we only evaluated it on MNIST and have not evaluated it on ImageNet, and we have performed a careful search over hyperparameters (e.g. layer choice, pooling size, reduced dimension, $k$, etc.), so we cannot conclude that their scheme will necessarily fail.

\section{Conclusion}

We show an attack on \knn and \knn-based classifiers and demonstrate that it improves on prior work.  Our experiments suggest that our attack can be used as a flexible and efficient method for evaluating the robustness of \knn and \knn-based models on real-world datasets.

\section{Acknowledgements}

This work was supported by the Hewlett Foundation through the Center for Long-Term Cybersecurity and by generous gifts from Huawei, Google, and the Berkeley Deep Drive project.

\bibliographystyle{IEEEtran}
\bibliography{advex,advml,ml,interpret,knn}

\end{document}